\documentclass[10pt, conference, letterpaper]{IEEEtran}

\usepackage{authblk}
\usepackage[ruled,vlined]{algorithm2e}
\usepackage{multicol}
\usepackage[utf8]{inputenc}
\usepackage{amsfonts}
\usepackage{amsmath}
\usepackage{amssymb}
\usepackage[T1]{fontenc}
\usepackage{microtype}
\usepackage[dvipsnames]{xcolor}
\usepackage{graphicx}
\usepackage{graphics}
\usepackage[font=footnotesize]{caption}
\usepackage{subcaption}
\usepackage{booktabs}
\usepackage{multirow}
\usepackage{cite}
\usepackage[export]{adjustbox}
\usepackage{breqn}
\usepackage{acronym}
\usepackage[keeplastbox]{flushend}
\usepackage{setspace}
\usepackage{bm}
\usepackage{stackengine}

\usepackage{smartdiagram}
\usepackage{tikz}
\usesmartdiagramlibrary{additions}

\usepackage{hyperref}
\definecolor{azure}     {rgb}{0,0.5,1}
\definecolor{dkpowder}  {rgb}{0,0.2,0.7}
\definecolor{deepred}   {rgb}{0.7,0,0}
\definecolor{deepblue}  {rgb}{0,0,0.7}
\definecolor{deepgreen} {rgb}{0,0.5,0}
\definecolor{deeporange}{rgb}{0.91, 0.41, 0.17}
\hypersetup{%
    pdfpagemode  = {UseOutlines},
    bookmarksopen,
    pdfstartview = {FitH},
    colorlinks,
    linkcolor = {dkpowder},
    citecolor = {dkpowder},
    urlcolor  = {dkpowder},
}

\usepackage{listings}

\lstdefinelanguage{BibTeX}{%
keywords={%
    @article,@book,@collectedbook,@conference,@electronic,@ieeetranbstctl,%
    @inbook,@incollectedbook,@incollection,@injournal,@inproceedings,%
    @manual,@mastersthesis,@misc,@patent,@periodical,@phdthesis,@preamble,%
    @proceedings,@standard,@string,@techreport,@unpublished%
},
comment=[l][\itshape]{@comment},
sensitive=false,
}

\DeclareMathOperator*{\argmax}{arg\,max}

\graphicspath{{/images}}
\makeatletter
\def\endthebibliography{%
  \def\@noitemerr{\@latex@warning{Empty `thebibliography' environment}}%
  \endlist}
\makeatother

\definecolor{label-running} {RGB}{ 31,119,180}
\definecolor{label-walking} {RGB}{255,127, 14}
\definecolor{label-jumping} {RGB}{ 44,160, 44}
\definecolor{label-standing}{RGB}{148,103,189}
\definecolor{label-sitting} {RGB}{140, 86, 75}
\definecolor{label-lying}   {RGB}{127,127,127}
\definecolor{label-falling} {RGB}{188,189, 34}
\definecolor{label-transit} {RGB}{ 23,190,207}

\IEEEoverridecommandlockouts

\title{\LARGE \bf Autonomous Blimp Control using Deep Reinforcement Learning }

\author{Yu Tang Liu$^{1,2}$, Eric Price$^{1,2}$, Pascal Goldschmid$^{2,1}$, Michael J. Black$^1$, Aamir Ahmad$^{2,1}$
\thanks{$^1$Max Planck Institute for Intelligent System, T{\"u}bingen, Germany.}
\thanks{ \{\tt\small firstname.lastname\}@tuebingen.mpg.de}
\thanks{$^2$Institute for Flight Mechanics and Controls, The Faculty of Aerospace Engineering and Geodesy, University of Stuttgart, Stuttgart, Germany.}
\thanks{ \{\tt\small firstname.lastname\}@ifr.uni-stuttgart.de}
        }

\begin{document}

\maketitle

\begin{abstract}

Aerial robot solutions are becoming ubiquitous for an increasing number of tasks. Among the various types of aerial robots, blimps are very well suited to perform long-duration tasks while being energy efficient, relatively silent and safe. To address the blimp navigation and control task, in our recent work \cite{Price:IAS:2021} we have developed a software-in-the-loop simulation and a PID-based controller for large blimps in the presence of wind disturbance. However, blimps have a deformable structure and their dynamics are inherently non-linear and time-delayed, often resulting in large trajectory tracking errors. Moreover, the buoyancy of a blimp is constantly changing due to changes in the ambient temperature and pressure. In the present paper, we explore a deep reinforcement learning (DRL) approach to address these issues. We train only in simulation, while keeping conditions as close as possible to the real-world scenario. We derive a compact state representation to reduce the training time and a discrete action space to enforce control smoothness. Our initial results in simulation show a significant potential of DRL in solving the blimp control task and robustness against moderate wind and parameter uncertainty. Extensive experiments are presented to study the robustness of our approach. We also openly provide the source code of our approach\footnote{https://github.com/robot-perception-group/AutonomousBlimpDRL}.

%
%
%
%
%
%
%
%
%
\end{abstract}

\begin{IEEEkeywords}
nonlinear control systems; reinforcement learning; blimp navigation; blimp control; aerial robotics; aerial vehicles
\end{IEEEkeywords}


\section{Introduction}
\label{sec:1_intro}


Autonomous unmanned aerial vehicles (UAVs) are becoming increasingly popular for various tasks, such as search and rescue, payload (medicine, food) delivery in difficult-to-reach areas, aerial cinematography and wildlife monitoring \cite{gonzalez2016unmanned, wang2016detecting,7139494, 4282475, aircap2019aerialswarms, mademlis2019high, markerlessmocap}. Current solutions rely on quadcopters and fixed-wings. Although quadcopters can hover in a fixed position, they are not able to accomplish long-term missions due to their short battery life. The situation is opposite for the fixed-wings, which have to move constantly to stay airborne. Therefore, for tasks involving long flight times, more payload and hovering over a small region, blimps provide an attractive solution.


A blimp is an airship without a rigid hull structure. Filled with helium, it becomes lighter than air and can hover for long periods without spending much energy. A blimp's weight is usually concentrated at its gondola, which creates a huge inertia to stabilize itself. From a control perspective, this makes the blimp an inherently stable plant \cite{li2007modeling} and allows it to recover from undesired states.


For blimp controller design, classic approaches usually rely on PID controllers \cite{1013654, 770044, 894672, takaya2006pid, 1626776} and nonlinear control \cite{1554402,1570450,4470489,1177111,LIU2020105610,Cheng2018,4776979, LIU2018231}. PID controller suffers from nonlinearity, and nonlinear control methods require a dynamic model of the system which is often difficult to acquire. Deep reinforcement learning (DRL), on the other hand, is a new control framework, which has achieved success in a variety of applications that present similar challenges \cite{NIPS2003_2455, Hwangbo_2017, Silver_2016, Mnih2013PlayingAW, zhu2018dexterous, bellemare2020autonomous}. The model-free RL approach is particularly useful when it is difficult to estimate physical parameters such as buoyancy and aerodynamic effects. The learning ability allows the controller to potentially adapt to the dynamic change caused by the environment. 



\begin{figure}[t]
    \centering
    \includegraphics[width=0.5\textwidth]{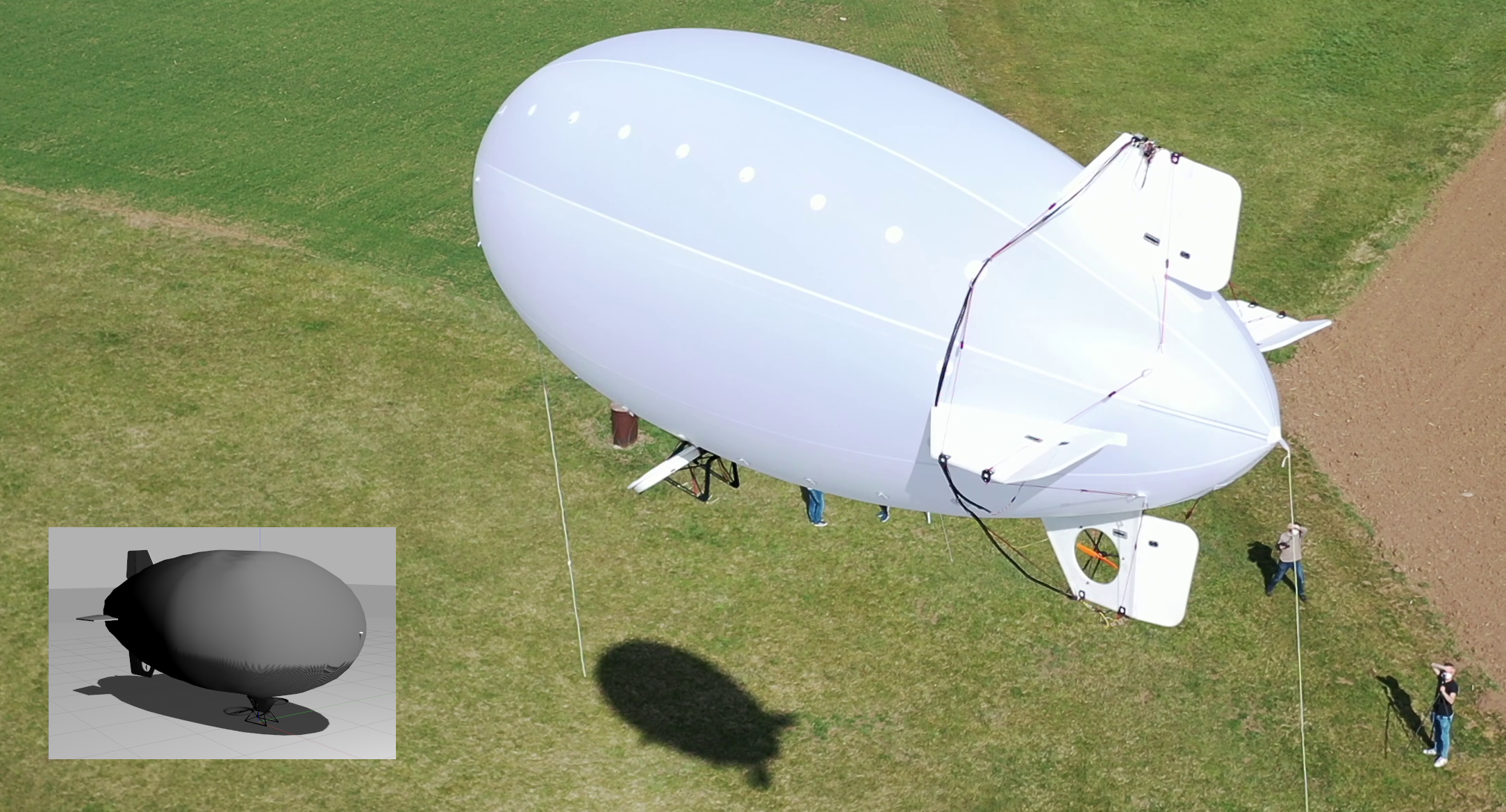}
    \caption{Our autonomous blimp during a flight. Unlike common designs, our blimp has thrust vectoring increasing its agility. Inset: a gazebo model of our blimp.}
    \label{fig:blimp_body}
\end{figure}

 In \cite{Price:IAS:2021} we developed a software-in-the-loop (SITL) simulation and a manually-tuned PID controller for the blimp control task. There, we demonstrated its ability to follow a waypoint sequence in the real world. To address the previously described issues of PID or nonlinear control approaches, in the current work we explore a DRL based method. Here, an RL agent, QRDQN\cite{mavrin2019distributional}, is deployed in the simulation environment to improve exploration and training stability. We design a variety of training task suites following the OpenAI-Gym framework \cite{stable-baselines3}, which is compatible with a variety of off-the-shelf RL platforms \cite{liang2018rllib, TFAgents,hoffman2020acme}. To achieve DRL training within a reasonable amount of time, we reduce problem difficulties by selecting task-specific features. We choose a discrete action space design to enforce continuity in the actuator commands. An action penalty is included in the reward function to regulate motor usage. To robustify the agent, we corrupt data by injecting noise to all observations and actions during training. We show that this integration address some of the common issues that may appear in a real world scenario.


\section{Related work}
\label{sec:2_related}

Control methods for blimps and airships, which have similar control schemes, have been well studied in the last decade \cite{5420403}. Classic approaches usually rely on \textbf{PID controllers}. Popular applications include visual servo\cite{1013654, 770044, 894672} and indoor miniature blimp\cite{takaya2006pid, 1626776}. The PID control class, while simple and robust, often suffers from plant nonlinearity. To overcome this weakness, advanced approaches have been developed using \textbf{nonlinear control theory}. Existing methods in this context include inverse optimal tracking control\cite{1554402}, dynamic inversion control\cite{1570450}, and the more investigated, backstepping control\cite{4470489,1177111,LIU2020105610}, robust control\cite{Cheng2018,LIU2018231}, and model predictive control \cite{4776979, LIU2018231}. However, these methods usually require a dynamic model which can be difficult to acquire.
The buoyancy of the blimp is heavily dependent on the constantly-changing surrounding environment. When temperature or pressure changes, buoyancy also changes and a controller has to adapt on the fly. Unfortunately, this effect has not been addressed in any of the prior works so far.

On the other hand, recently there has been a surge of interest in applying RL to robotics \cite{singh2021reinforcement}. The earliest attempts in the \textbf{classic RL} use Gaussian processes (GPs) for system identification \cite{4209179} and policy learning\cite{5152660, 4399531}. Despite sample efficiency, GPs are hard to scale up with problem dimensions and demand higher computational resources. As a result, they are able to achieve success only on low dimensional tasks, such as 1-D altitude control. \textbf{DRL}, on the other hand, leverages NNs for policy approximation and has achieved much success. This policy class can interpret rich representations and derive diverse behavior. For example, Nie et al.\ \cite{nie2019three} train two DQN agents for rudder and elevator control of a blimp, respectively, and demonstrate a better performance than a PID controller. In the field of autonomous underwater vehicle (AUVs)\footnote{Due to the lack of existent work, we include AUVs but only focus on those that have a fairly similar shape and task specification}, Carlucho et al. \cite{8604791} use a Nessie-VII model with continuous action/observation space using DDPG. 

The main challenge with DRL is the lack of sample efficiency. In order to scale up the DRL formulation with the problem dimension, a highly increased amount of environment interactions is needed by the agent. Other challenges include adapting a trained policy to real-world scenario \cite{zhang2019bridging}, action smoothness\cite{caps2021}, etc. Furthermore, issues such as partial observability, disturbances and noise could also lead to unexpected behavior. Issuing stability certificates to RL agents is also an ongoing research topic. In case of AUVs, the robustness issue is addressed in \cite{9414937}, by training a PPO agent in an adversarial fashion at the cost of conservative behavior. 


To increase sample efficiency and finish training within a reasonable time budget, in this work we train the policy network with a value-based RL agent, QRDQN, which is a more stable variant in the DQN family. Action and observation space are injected with noise during training to robustify the agent. Lastly, to enforce actuator smoothness, we choose a discrete action space (Sec.\ref{sec:3_method_task_formulation_observation_and_action_space}).


\section{Methodology}
\label{sec:3_method}

\subsection{Preliminary}
Blimp shape and architecture can vary a lot from one another. Therefore, we first describe our blimp, which has 8 actuators (but our approach is agnostic to different configurations). The two main motors (for thrust) are attached to a servo which allows thrust vectoring. At the tail of the blimp, there are four fins controlling yaw and pitch angle and a tail motor, attached to the bottom fin, generating horizontal thrust allowing further yaw controllability. The state of the actuators can be denoted as 

\begin{equation}
  \label{eqn:3_method_action_state_representation}
  \begin{gathered}[l]
    s_t^{act}=(m_{(0:2)}, s, f_{(0:3)})_t\in\mathbb{R}^{8},
  \end{gathered}
\end{equation}

where $m_{0:2},s,f_{0:3}$ stand for motor, servo, and fin states, respectively. Our goal is to navigate this blimp to any given waypoint in the space by controlling these actuators.


\subsection{Formulation}
\label{sec:3_method_task_formulation}

\subsubsection{Control}
We formulate the problem as a path following task as seen in previous works\cite{1626776, nie2019three, 5611169}. In this setting, an imaginative path reference is generated based on waypoints for the controller to follow. Casting the path following task as a DRL problem, in this section we show how we reduce the state space size and maintain a reasonable training time. Since the blimp does not have a lateral movement control, we only need to consider longitudinal and altitude control. This allows us to easily decompose the problem into a planar navigation control task and an altitude control task. The objective of the planar navigation control is to control the blimp to arrive at any waypoint in the xy-plane whereas the altitude control is to reach the desired altitude of the waypoint. 

\begin{figure}[t]
    \centering
    \includegraphics[width=0.5\textwidth]{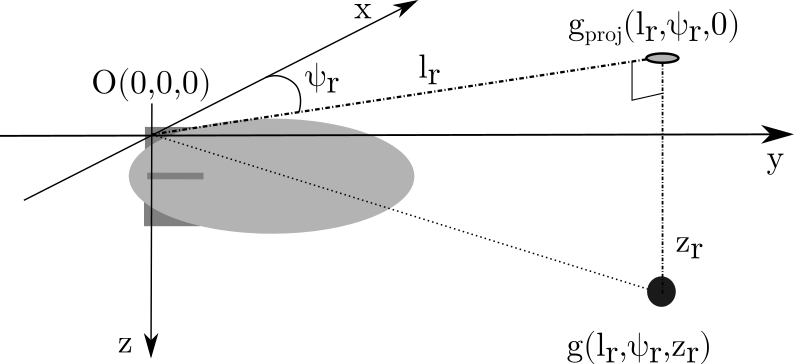}
    \caption{The bodyframe in NED cylindrical coordinate system. $O$ is at the top fin of the blimp where GPU and IMU sensors are mounted. $g_{proj}$ is the projection on xy-plane of the waypoint $g$.}
    \label{fig:blimp_diagram}
\end{figure}

Given the blimp position at $O=(0,0,0)$ and a target waypoint at $g=(l_r,\psi_r, z_r)$ in body frame cylindrical coordinates (Fig.~\ref{fig:blimp_diagram}), the control objective of the planar navigation control is the minimization of the relevant distance and yaw angle, or $\min_{a\in A}(|l_r|, |\psi_r|)$. The objective of the altitude control is to minimize the relevant altitude, or $\min_{a\in A}{|z_r|}$. The spatial information between the target and the blimp can be fully contained in $g=(l_r, \psi_r, z_r)$. Although it is possible to train the DRL method only using $g$, this minimal setting ignores the velocity and pitch state of the blimp, leading to instability in training and an uncontrolled behavior when reaching the waypoint. 

We denote the velocity of the blimp as $V=(u,v,w)$, and attitude (roll, pitch, yaw) as $\Phi=(\phi,\theta,\psi)$. Assuming near zero lateral movement in the blimp (i.e. $v,\phi\simeq0$), the velocity and pitch angle information can be encoded by velocity magnitude ($|V| = ||(u,v,w)||_{2}$) and the altitude velocity ($w = |V|sin\theta$), alone. We augment our state with this velocity information and derive a compact representation for the overall state as

\begin{equation}
  \label{eqn:3_method_state_representation}
  \begin{gathered}[l]
    s_t^{blimp}= (l_r, \psi_r, z_r, |V|, w)_t,
  \end{gathered}
\end{equation}which encodes all the spatial, velocity, and attitude information. As we do not directly use the pitch angle information, this representation is agnostic to sensor calibration error in pitch angle, which can be easily misaligned from the simulation.

\subsubsection{Markov Decision Process}
We consider the RL problem as an infinite horizon discrete time Markov Decision Process, $M$, defined by a tuple $(S,A,P,R,\gamma)$ \cite{sutton2018reinforcement}. At any time step $t{\small \in} \mathbb{R}^+$ and state $s_t {\small \in} \mathbb{R}^S$, an agent draws an action $a_t$ from a discrete action space $ A=\{0,1,...,K\}$ given the policy distribution $a_t {\small \sim} \pi_{\theta}(\cdot|s_t)$ parameterized by $\theta$. The environment then samples the next state from an unknown transition distribution, i.e. $s_{t+1} {\small \sim} P(\cdot|s_t, a_t)$. A reward is received based on some reward function $r_t=R(s_t,a_t)$. Given the discount factor $\gamma {\small \in} [0,1)$, the goal of the agent is to find the optimal policy parameter $\theta$ that comes with the highest cumulative discounted reward (\ref{eqn:total_sum_of_discounted_reward}),

\begin{equation}
  \label{eqn:total_sum_of_discounted_reward}
  \begin{gathered}
    \pi^* = \argmax_{ \pi_{\theta} } \mathop{\mathbb{E}}_{ \pi }[ \sum_{t}^{\infty} \gamma^t r_t | a_t {\small \sim} \pi(\cdot|s_t), s_{t+1} {\small \sim} P(\cdot|s_t,a_t) ]
  \end{gathered}
\end{equation}

\subsubsection{Observation and Action Space}
\label{sec:3_method_task_formulation_observation_and_action_space}
The full actuator state, $s_t^{act}$, is described in (\ref{eqn:3_method_action_state_representation}). Since we do not allow differential thrust, symmetric actuators are always in the same state. Thus, we only need to feedback one of them. The reduced state of actuators is therefore defined as $s_t^{act'}=(m_{(0,2)}, s, f_{(0,2)})_t\in\mathbb{R}^{5}$. The full state $s_t$ for the DRL formulation, as used in (\ref{eqn:total_sum_of_discounted_reward}), is now obtained below as the concatenation of $s_t^{blimp}$ and $s_t^{act'}$.

\begin{equation}
  \label{eqn:3_method_full_state_representation}
  \begin{gathered}[l]
    s_t=(s_t^{act'}, s_t^{blimp}).
  \end{gathered}
\end{equation}

Note that all states are scaled to the range $[-1,1]$ and zero-initialized. 
To prevent significant and sudden changes in the actuator command, we use discrete action space, denoted as $a_t\in A=\{0,1,...,K\}$. The action command is then mapped to the actuator command $\delta s_{a_t} {\small \in} \mathbb{R}^{8}$ following Table~\ref{tab:3_method_navigation_discrete_meta_act} and then summed up with the actuator state. This process is described below in (\ref{eqn:3_method_action_selection_process}).

\begin{equation}
  \label{eqn:3_method_action_selection_process}
  \begin{gathered}
    a_t \sim \pi(\cdot|s_t) \\
    \delta s_{a_t} = Table~\ref{tab:3_method_navigation_discrete_meta_act}(a_t) \\
    s_{a_t} \leftarrow s_{a_t} + \delta s_{a_t} \\
  \end{gathered}
\end{equation}

\subsubsection{Reward Function}
\label{sec:3_method_task_formulation_reward_function}

The control tasks of the UAVs usually involve navigation and hover. Navigation requires moving the robot in space by specifying a target position or following a sequence of targets,  whereas hovering requires staying near the target position. These two tasks can be combined and trained with the same setup by using appropriate reward functions. When the goal is far away, we use a reward function for navigation only, otherwise a hover reward function. The reward function is defined by (\ref{eqn:3_method_reward_function})

\begin{equation}
  \label{eqn:3_method_reward_function}
  \begin{gathered}[l]
    r_t = w_{0}r_t^{success}+ w_{1}r_t^{track} + w_{2}r_t^{act},
  \end{gathered}
\end{equation}where $w_{0:2}=(1,0.95,0.05)$ in this paper. The agent receives a success reward, $r_t^{success}$, if the task is completed. Tracking reward, $r_t^{track}$, indicates the tracking performance. Action reward, $r_t^{act}$, is defined to regularize actuator commands.

\begin{equation}
    \label{eqn:3_method_success_reward_function}
    \begin{gathered}[l]
        r_t^{success} = \begin{cases}
        1 &\text{if $d(s_{blimp}, s_{target}) \leq \epsilon$ } \\
        0 &\text{otherwise}
        \end{cases},
    \end{gathered}
\end{equation}

\begin{equation}
    \label{eqn:3_method_switch_reward_function}
    \begin{gathered}[l]
        r_t^{track} = \begin{cases}
        r_t^{hover} &\text{if $d(s_{blimp}, s_{target}) \leq \epsilon$ } \\
        r_t^{navigate} &\text{otherwise}
        \end{cases},
    \end{gathered}
\end{equation}where $d(s_{blimp},s_{target})$ measures euclidean distance between the blimp and the target position. (\ref{eqn:3_method_success_reward_function}-\ref{eqn:3_method_switch_reward_function}) indicate if this distance is short enough, the reward will be switched from navigation reward, $r_t^{navigate}$, to hover reward, $r_t^{hover}$, which does not take yaw component into account (\ref{eqn:3_method_navigate_reward_function}-\ref{eqn:3_method_hover_reward_function}). Note that we could also use $|V|$ or $w$ in the reward function to address other tasks.

\begin{align}
    &r_t^{navigate} = -i_0|z_r| - i_1|l_r| - i_2|\psi_r|, \label{eqn:3_method_navigate_reward_function}\\
    &r_t^{hover} = -j_0|z_r| - j_1|l_r|, \label{eqn:3_method_hover_reward_function}\\
    &r_t^{act} = -k_0||m_0,m_1,m_2||_2,
\end{align}where $i_{0:2}=(0.1,0.7,0.2),j_{0:1}=(0.3,0.7),k_{0}=1$ in this paper.

\begin{table}[h!]
    \centering
    \resizebox{0.465\textwidth}{!}{%
        \begin{tabular}{ |p{0.2cm}|p{2.1cm}||p{5.6cm}|  }
         \hline
             \textit{A} & Name & $\delta s_{a_t}=\delta [m_2, f_0, f_1, f_2, f_3, s, m_0, m_1]_t$\\
             \hline
             0 & IDLE & [0, 0, 0, 0, 0, 0, 0, 0] \\
             1 & THRUST+ & [0, 0, 0, 0, 0, 0, 0.01, 0.01] \\
             2 & THRUST- & [0, 0, 0, 0, 0, 0, -0.01, -0.01] \\
             3 & NOSE_UP & [0, 0.025, 0.025, 0, 0, 0, 0, 0] \\
             4 & NOSE_DOWN & [0, -0.025, -0.025, 0, 0, 0, 0, 0] \\
             5 & NOSE_LEFT & [0.025, 0, 0, 0.025, 0.025, 0, 0, 0] \\
             6 & NOSE_RIGHT & [-0.025, 0, 0, -0.025, -0.025, 0, 0, 0] \\
             \hline
        \end{tabular}
    }
    \caption{discrete action space $\delta s_{a_t}$: ($\%$). The notation $f_{0:3}$ correspond to the angle of left/right/top/bottom fins. Note that in this work thrust vectoring is disabled ($s,\delta s=0$).}
    \label{tab:3_method_navigation_discrete_meta_act}
\end{table}

\subsection{Training Setup}
\label{sec:3_method_policy_training}
In this section, we describe the important factors that contribute to stabilize the training and increase the robustness of the trained policy. During training, the target position is sampled randomly within the range of $200$ cubic meters w.r.t.\ the blimp. Random sampling is important to increase sample diversity and avoid overfitting to a specific track. We reset the task only after $200$ seconds so that there is sufficient time for the blimp to reach any target, and use the spare time to learn to stay within the target range. During training, to increase the robustness of the policy, observations and actions are injected with $5\%$ of noise and clip to the range $[-1,1]$. Lastly, while the simulation step time is $0.1s$, the policy step is $0.5s$. Since the blimp has a relatively long response time, we found it important to increase the step time for the action to take effect.

\begin{figure}[t]
   \centering
   \includegraphics[width=0.5\textwidth]{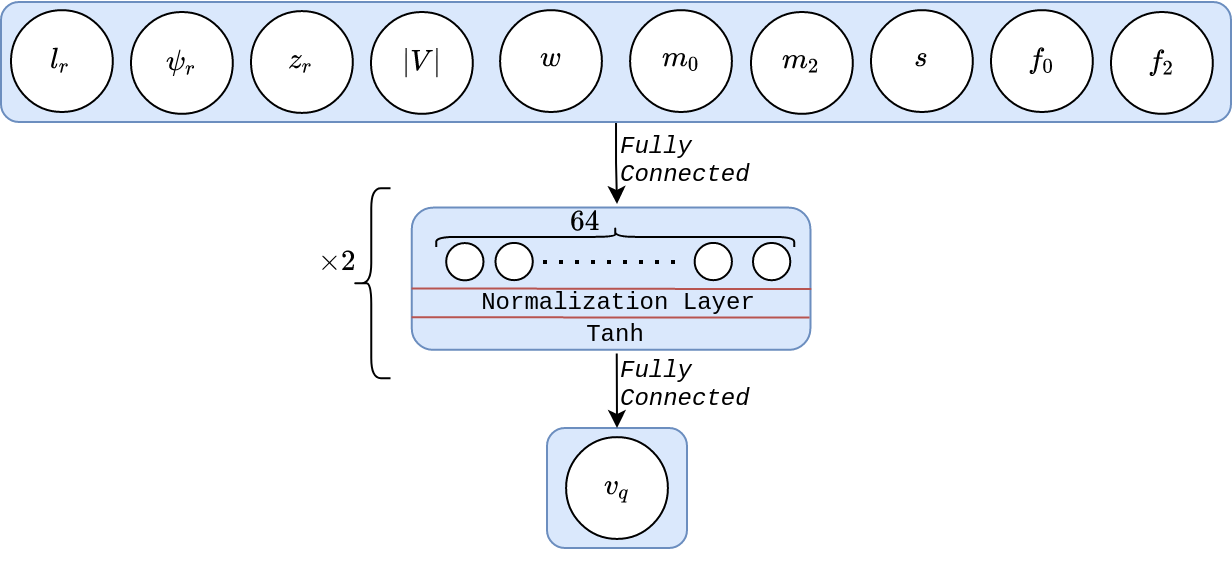}
  \caption{The policy networks has the weights of $64$ neurons by $2$ layers. To prevent vanishing/exploding gradient, we add normalization layers to every linear layers. $v_q$ is a value array of size $K$ which evaluates each possible actions in $A$. Policy chooses action based on a greedy law, $a_t=argmax(v_q)$.}
   \label{fig:nn_connect}
\end{figure}

We train the policy network with the QRDQN agent. The value-based method is in general more sample efficient compared to gradient-based methods and can therefore accelerate the training. QRDQN leverages a quantile network to estimate value function, which is important to stabilize training by alleviating chattering \cite{mavrin2019distributional} effect and extreme value estimates. The architecture of our 
policy network is shown in Fig.~\ref{fig:nn_connect}. To reduce training time, we only apply less than $13$ quantiles and sacrifice some estimation resolution. 

\section{Experiment}
\label{sec:4_experiment}
In the experiments, with the real world scenario in mind, we address the following questions. Is our DRL formulation with the compact representation able to solve the complex 3-D path-following task?  In order to answer this question, we introduce the navigation and hovering tasks which are the building blocks for further complicated tasks. To evaluate the agent performance, a PID controller (from our previous work \cite{Price:IAS:2021}) is considered as a benchmark, which is simple but well-known for its robustness. Finally, through various experiments (see sub-sec.~\ref{subsec:robustness_study}) we evaluate if the RL agent is ready to be deployed in the real world. In other words, we evaluate the agent's robustness against unknown environmental changes.

\subsection{Experiment Setup}
We integrate our RL training environment in the ROS/Gazebo SITL simulation following the OpenAI-Gym framework. The QRDQN implementation is based on the StableBaseline3 \cite{stable-baselines3}. The agent is trained for 7 days on a single computer (AMD Ryzen Threadripper 3960X, 24x 3.8GHz, NVIDIA GeForce RTX 2080 Ti, 11GB). Our simulation environment is designed based on our real robotic blimp (see Fig.\ref{fig:blimp_body}). The baseline PID controller is well-tuned to the simulation environment. Our previous work \cite{Price:IAS:2021}) has shown that we could deploy it to the real world without further tuning, which implies a good quality of the simulation. 

\subsection{Task Suite}
To evaluate the performance of the agent, the navigation and hover tasks are introduced in the Sec.\ref{sec:3_method_task_formulation}.  For convenience, we visualize the target waypoints in the world ENU frame. 

\subsubsection{\textbf{Navigation}} 
\begin{figure}[h!]
    \centering
    \begin{subfigure}[b]{0.5\textwidth}
        \centering
        \includegraphics[width=\textwidth]{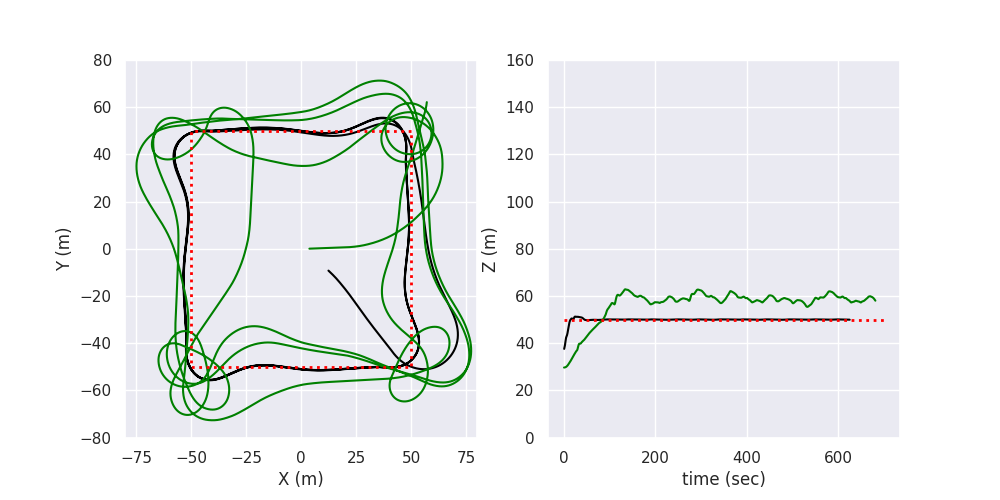}
        \caption{Navigation Task: Left -- the planar trajectory of the blimp. Right -- the altitude trajectory. Red: reference. Black: PID controller. Green: RL policy. The PID controller completes the task around $10\%$ faster than the RL policy, which seems to favor an altitude $7$ meters above the target altitude.}
        \label{fig:square}
    \end{subfigure}

    \begin{subfigure}[b]{0.5\textwidth}
        \centering
        \includegraphics[width=\textwidth]{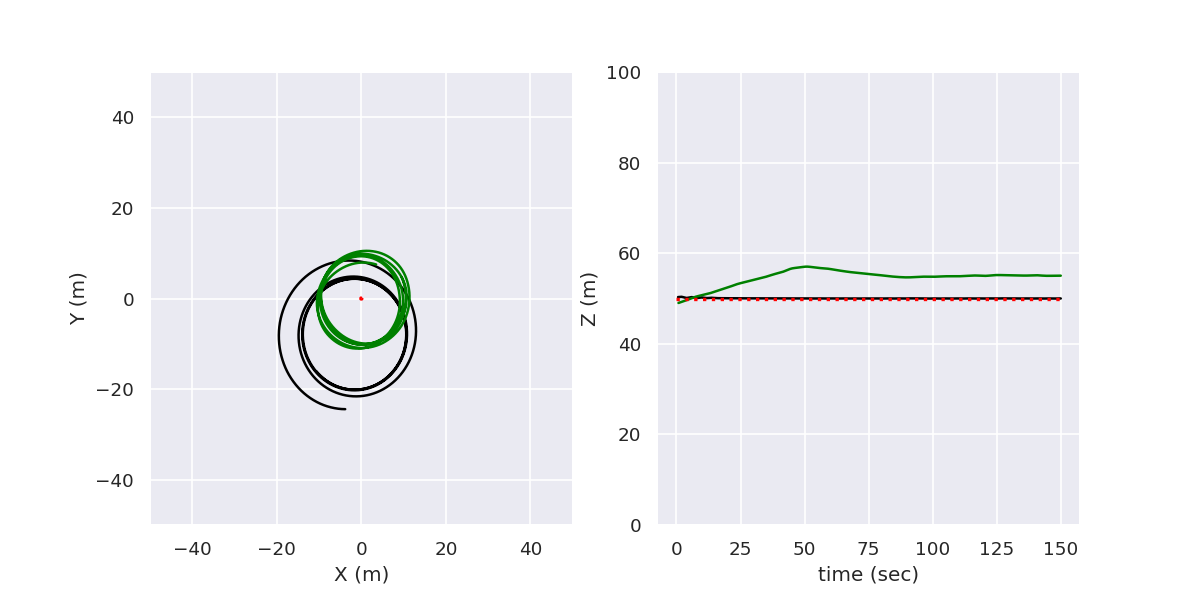}
        \caption{Hover Task: Left: The planar trajectory of the blimp. Right: The altitude trajectory. Red: reference. Black: PID controller. Green: RL policy. The RL policy hovers with less radius around the target but it loiters $5$ meters above the target altitude.}
        \label{fig:hover}
    \end{subfigure}

    \caption{Comparison of PID controller and RL policy in navigation and hover tasks. }
\end{figure}

Four waypoints are created at an altitude of $50$ m to form a square with sides of $100$ m each. This has to be traversed in a counter-clockwise direction. A waypoint is registered when the blimp is within $15$ m radius and then the next waypoint is triggered. The early waypoint trigger allows less overshoot and achieves better performance. To make sure the comparison is fair, the track has to be performed 3 times to be marked as complete. The velocity for the PID controller is set to have a slow reference speed of $2m/s$ to prevent overshoot, while the agent is not subjected to any speed limit but $50\%$ maximum throttle. The results are shown in Fig.~\ref{fig:square}.
\
The PID controller has a stable performance during the whole task and remains a challenging baseline. On the other hand, although our trained RL policy can complete the navigation task successfully, it shows higher discrepancy from the reference path. It spends most of the time hovering above the waypoints and reduces the altitude until the next waypoint gets triggered.

\subsubsection{\textbf{Hovering}}
The hovering task requires the blimp to stay as close to the target as possible without spending excessive amount of energy. The target waypoint is positioned at $(0,0,50)$ in the world ENU frame. The blimp is spawned at the target position.  The result is in Fig.~\ref{fig:hover}. The PID controller requires a larger radius compare to the RL policy. On the other hand, similar to the navigation task, the RL policy tends to hover $5$ meters above the target altitude. Our initial reasoning for this behavior was lack of training. However, after continuing training the same policy, the results become worse. When the waypoint is an arbitrary point in space and far from the origin, the agent accelerates towards the target at first, then hovers close to it, and finally makes a slow approach towards it. We argue that hovering close to the target altitude gives long-term advantages to the agent, analyzed as follows. First, the agent receives more action rewards ($r_t^{act}$) as it does not need to command anymore but only needs to wait until it slowly approaches the target altitude. Second, during this time, the distance is short enough to receive a good amount of hover reward ($r_t^{hover}$); and if the agent would rush to the target, it is most likely to overshoot and spend an excessive amount of energy to come back to the hover position, and subsequently overshoot again. Third, since the speed of the blimp becomes very small when approaching the target, the agent can easily stay longer within the target range and continuously receive abundant success rewards ($r_t^{success}$). 

\begin{figure}[h!]
    \centering
    \begin{subfigure}[b]{0.5\textwidth}
        \centering
        \includegraphics[width=\textwidth]{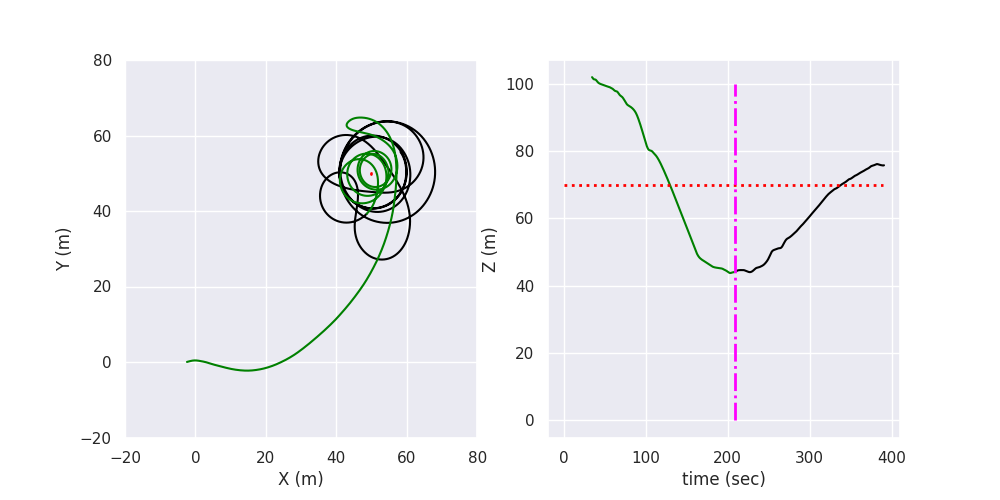}
        \caption{What happens if the agent is spawned far from the target? Left: the planar view of the blimp approaching the target at (50,50,70). Right: the altitude trajectory of the blimp. Red: waypoint. The blimp significantly loses its altitude near the target position.}
        \label{fig:attack}
    \end{subfigure}
    \begin{subfigure}[b]{0.5\textwidth}
        \centering
        \includegraphics[width=\textwidth]{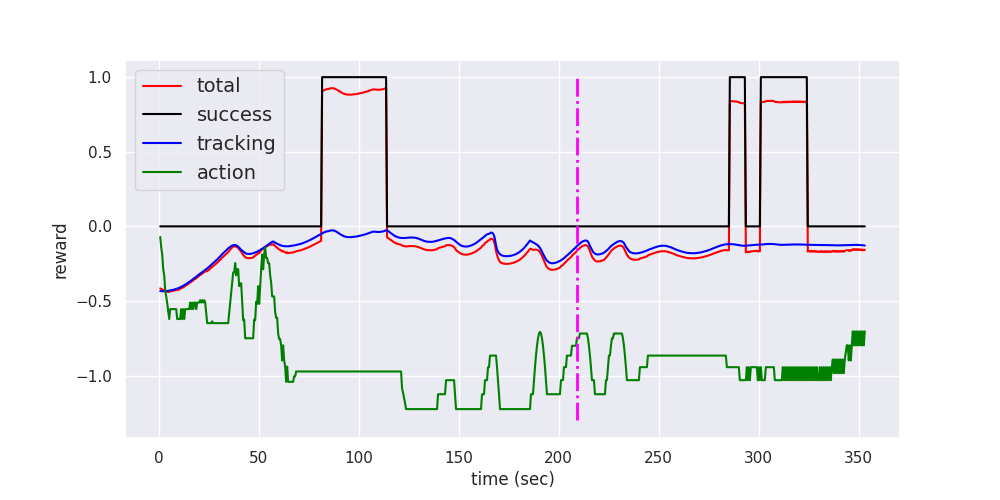}
        \caption{Reward exploitation by the agent. Red: total reward. Black: Success reward. Blue: tracking reward. Green: action reward. The agent still receives a good amount of reward despite the altitude loss.}
        \label{fig:attack_reward}
    \end{subfigure}
    \caption{Analysis of the hovering behavior. Magenta: the time agent responds to altitude loss.  The total reward is calculated based on the (\ref{eqn:3_method_reward_function}). The tracking reward does not penalize altitude discrepancy enough and causes the strange behavior.} 
    \label{fig:weird}
\end{figure}

We are able to reproduce this behavior as shown in Fig.~\ref{fig:attack}. The blimp is spawned at $(0,0,100)$, which is above the target altitude $(50,50,70)$. We first observe that the blimp approaches the target, then stays close to it with a low speed. During this time, it still receives a good amount of tracking and success reward as shown in Fig.~\ref{fig:attack_reward}. The blimp then continues to sink $25$m below the target, after which the policy brings it back and raises it above the target altitude. This overall behavior required $\sim 200$s. This is followed by the hovering behavior as observed in Fig.~\ref{fig:hover}. Such a behavior comes from the fact that the total reward is dominated by the success reward and the loss of altitude does not result in significant punishment in tracking reward. Therefore, manipulating the reward function and increase altitude weight could help get rid of this behavior. 

\subsection{Robustness Study}
\label{subsec:robustness_study}
\begin{figure}[h!]
        \centering
        \includegraphics[width=0.5\textwidth]{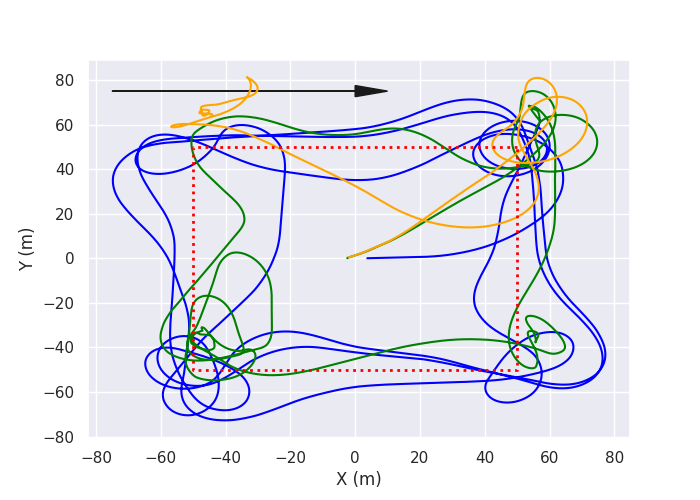}
        \caption{Effect of wind. Blue: no wind. Green: $2m/s$ wind. Orange: $4m/s$ wind. The dark arrow indicates the direction of the wind field. This experiment was conducted for 3 laps for the `no wind' case but for only 1 lap for the other two cases. When the wind speed is $4m/s$ the agent is not able to complete the task.}
        \label{fig:wind}
\end{figure}

\begin{figure*}[h]
	\includegraphics[width=\textwidth]{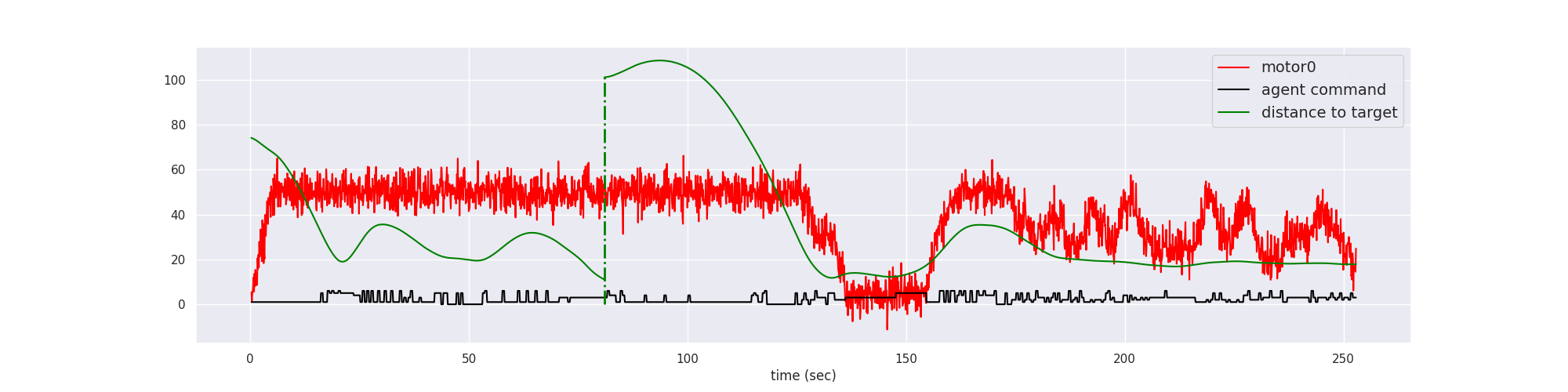}
	\caption{Behavior of the DRL policy when the blimp is subject to $4m/s$ wind: Motor0 is the motor output $(\%)$. Distance to target is calculated by euclidean distance $(m)$. The left side of the green dotted line is the first waypoint, and right side is the second. The second waypoint is harder to reach since the blimp has to move against the wind. Whenever it is close to the target, it starts to transition into hover mode. This causes the blimp to be blown away by the wind, from which it is never able to recover. }
	\label{fig:wind_action}
\end{figure*}

To show the robustness of the agent, we test our agent with i) a fixed wind field, ii) changes in the blimp buoyancy, and ii) changes in the weight distribution along the gondola. In Fig.~\ref{fig:wind}, the agent is able to handle small wind disturbance at $2m/s$ but fails at $4m/s$. Under the wind condition, it takes a significantly large amount of time to finish the task. Notice that when the wind is at $2m/s$, the agent trajectory seems to be smoother. This is because wind slows down the agent and prevents overshooting the target. At the wind speed of $4m/s$, the agent tends to slow down when it approaches the target as in Fig.~\ref{fig:wind_action}. This is the side effect from training navigation and hovering task together as the agent tries to slow down to stay within the range of success reward. It starts to slow down around $40$ m to the target. As a result, although the agent has enough thrust power to overcome $4m/s$ wind, it gradually reduces both motors to zero speed and then is blown away by the wind. A na\"ive workaround is to toggle the target switch when the blimp is $40$ m from the target. But a toggle with such a huge radius is not realistic. Notice that in Fig.~\ref{fig:wind_action}, the motors always have smooth transition due to the discrete action design which only allows $1\%$ motor speed change every $0.5$ seconds.

\begin{table}[h!]
	\begin{subtable}[t]{0.24\textwidth}
		\centering
		\resizebox{0.7\textwidth}{!}{%
			\begin{tabular}{ | p{1.5cm} | p{1.5cm} | }
				\hline
				\textit{Buoyancy} & Avg. Time (sec) \\
				\hline
				100\% & 238 \\
				95\%  &  545\\
				90\% & NA \\
				85\% & NA \\
				80\%  & NA  \\
				\hline
			\end{tabular}
		}
		\caption{Effect of buoyancy on average time to complete a square.}
		\label{tab:buoyancy}
	\end{subtable}
	\hfill
	\begin{subtable}[t]{0.24\textwidth}
		\centering
		\resizebox{0.7\textwidth}{!}{%
			\begin{tabular}{ | p{1.5cm} | p{1.5cm} | }
				\hline
				\textit{Added mass (g)} & Avg. Time (sec) \\
				\hline
				0       & 238 \\
				-100 &  237\\
				-250 & NA \\
				100 & 328 \\
				250 & NA  \\
				\hline
			\end{tabular}
		}
		\caption{Effect of trim weight on average time to complete a square}
		\label{tab:weights}
	\end{subtable}
	\caption{Buoyancy and weights change cause significant impact on the agent}
	\label{tab:vertical_changes}
\end{table}

Another common scenario for the blimp in real world is the buoyancy change. Depending on the weather, the buoyancy of the blimp can change significantly. We test the performance of the agent with a decreasing amount of buoyancy w.r.t.\ the original state. Not to our surprise, the result in Tab.~\ref{tab:buoyancy} shows that the decreasing amount of buoyancy does lead to worse performance. With $95\%$ buoyancy, policy performance suffers significantly and takes much longer to finish the task. Lower than that the agent is not able to control the blimp at all. The effect of weight distribution (also commonly affected in real world) also can not be ignored as it could introduce unnecessary vibrations if not balanced. To this end we perform another experiment, where we add and remove ballast to the front end of the blimp to break this balance. Results in Tab.~\ref{tab:weights} suggest that $100$g of mass change does not affect the performance, but larger than that would impair the policy. These two experiments have shown that the RL agent is currently sensitive to the environment changes. We expect to improve the performance of the agent by increasing the penalty for the altitude loss.




\section{Discussion}
\label{sec:5_discussion}
In this work, we integrated the ROS/Gazebo SITL blimp simulation together with the RL training environment. We have derived a compact representation of the state space and action space which allows less training time and guarantee the actuator continuity. We have shown that such a setting is able to successfully complete the task. The trained policy network has a certain degree of robustness against wind and parameter uncertainty. 

On the other hand, we have observed and analysed how the agent exploits the reward function. 
The altitude loss is unacceptably large for this agent to be deployed to the real world. Increasing the altitude reward weights and punishing the altitude loss could potentially address this issue. However, further experiments are needed to verify this hypothesis. In this work, the reverse thrust and thrust vectoring were not enabled. Given a more diverse action space, the agent is more likely to gain more rewards by staying closer to the target. Another problem is that when training navigation and hover in the same time the agent learns the conservative behavior when approaching a waypoint. This could be potentially eliminated by including disturbance to the training and making it harder to exploit the weakness of this approach. A more promising solution would be multi-task learning which trains navigation and hovering task independently. 

There are many other open issues not been addressed in this work so far. In real world experiments, not presented in this paper, we have encountered several difficulties even when flying with a PID controller. For example, in this work we assumed the lateral movement can be neglected and longitude velocity is always positive. To our observation in real world, this is a dangerous assumption as it does not hold in the presence of moderate to strong wind. When the wind speed is larger than the vehicle's, the speed can become negative and lateral movement can be created if the wind is blowing from the side. This can be dangerous and cause undesired behavior for the policy network. 

Finally, blimp control has not received enough attention and still remains an underdeveloped field. RL-based methods do not provide any stability guarantee but provide the potential to learn continuously from data and improve its own performance. Conversely, the nonlinear controllers are robust against parameter uncertainty and disturbance at the expense of control performance. How to leverage these two approaches is the key to the success of future blimp control methods. Since blimp dynamic is heavily dependent on the environment, it serves as a perfect robotic platform to study adaptive learning control. Secondly, the modern DRL algorithms are still not sample efficient enough. Our next step is to leverage parallel training to accelerate gathering experience. This also allows us to increase the diversity in the training and offer the potential to leverage multi-tasking learning as mentioned in \cite{espeholt2018impala}. Lastly, for the agent to counter partial observations such as wind disturbances, it is important to include past experiences in the decision-making process. For example, a recurrent network architecture might be a possible solution.



\bibliographystyle{IEEEtran}
\bibliography{biblio}
\end{document}